\documentclass[10pt,doublecolumn]{IEEEtran}
\usepackage{amsmath,amsthm,amsfonts,amssymb,bm,mathrsfs}
\usepackage{graphicx,subfigure}
\usepackage{cite}
\usepackage{algorithm}
\usepackage{algorithmic}
\usepackage{multirow,booktabs,threeparttable}
\usepackage{tablefootnote}
\usepackage{color}
\usepackage{verbatim}
\graphicspath{{fig/}}

\theoremstyle{remark}

\theoremstyle{plain}

\usepackage{algorithm}
\usepackage{algorithmic}

\begin{document}
\title{Brain-Inspired Stigmergy Learning}

\author{\IEEEauthorblockN{Xing Hsu, Zhifeng Zhao, Rongpeng Li, Honggang Zhang}\\
	College of Information Science and Electronic Engineering\\
	Zhejiang University, Zheda Road 38, Hangzhou 310027, China\\
	\IEEEauthorblockN{Emails: $\{$hsuxing, zhaozf, lirongpeng, honggangzhang$\}$@zju.edu.cn } \\
}

\maketitle
\thispagestyle{empty}
\begin{abstract}
Stigmergy has proved its great superiority in terms of distributed control, robustness and adaptability, thus being regarded as an ideal solution for large-scale swarm control problems. Based on new discoveries on astrocytes in regulating synaptic transmission in the brain, this paper has mapped stigmergy mechanism into the interaction between synapses and investigated its characteristics and advantages. Particularly, we have divided the interaction between synapses which are not directly connected into three phases and proposed a stigmergic learning model. In this model, the state change of a stigmergy agent will expand its influence to affect the states of others. The strength of the interaction is determined by the level of neural activity as well as the distance between stigmergy agents. Inspired by the morphological and functional changes in astrocytes during environmental enrichment, it is likely that the regulation of distance between stigmergy agents plays a critical role in the stigmergy learning process. Simulation results have verified its importance and indicated that the well-regulated distance between stigmergy agents can help to obtain stigmergy learning gain.  
\end{abstract}

\begin{IEEEkeywords}
Stigmergy, Astrocytes, Synapses, Calcium Waves, Neural Networks, Artificial Intelligence, Machine Learning
\end{IEEEkeywords}

\section{Introduction}
Stigmergy was first introduced by French entomologist Pierre-Paul Grass\`e in 1950s\cite{Grass1959La}\cite{U1985Grass} when studying the behavior of social insects. The word stigmergy is a combination of the Greek words ``stigma" (outstanding sign) and ``ergon" (work), indicating that some activities of agents are triggered by external signs, which themselves may be generated by agent activities\cite{Informatik2000Return}. Stigmergy allowed Grass\`e to explain why insects of very limited intelligence, without apparent communications, can collaboratively tackle complex tasks, such as building a nest. Another definition given by Heylighen is\cite{Heylighen2011Stigmergy}: stigmergy is an indirect, mediated mechanism of coordination between actions in which a perceived effect of an action stimulates the performance of a subsequent action.

Stigmergy has been widely studied in the behavior of social insects\cite{Dorigo1999Ant}\cite{Kassabalidis2001Swarm}. But this concept is rarely mentioned in the brain, which has been regarded as the most complex system so far. As important glial cells in brain's Central Nervous System (CNS), astrocytes are traditionally placed in a subservient position, which supports the physiology of the associated neurons. However, recent experimental neuroscience evidences indicate that astrocytes also interact closely with neurons and participate in the regulation of synaptic neurotransmission\cite{Haydon2006Astrocyte}. These evidences have motivated new perspectives for the research of stigmergy in the brain.

There is a large number of complicated biochemical reactions between synapses and astrocytes to support the implementation of various brain functions. Basically, each astrocyte contains hundreds or thousands of branch microdomains, and each of them encloses a synapse\cite{Araque2014Gliotransmitters}, as illustrated in Fig. 1. The synaptic activity will elevate the concentration of $\rm{Ca^{2+}}$ in the corresponding microdomain. $\rm{Ca^{2+}}$ and inositol-1,4,5-triphospate ($\rm{IP_3}$), as important messengers within astrocytes, are believed to expand the influence of synaptic activities\cite{Mesiti2017Astrocyte}. The range of influence is determined by the level of synaptic activity as well as the distance between coupled branch microdomains. Besides, these branch microdomains with the elevated concentration of $\rm{Ca^{2+}}$ will provide an adjustment for the wrapped synapses. Therefore, as explained in Section II, general stigmergy can be mapped into this neural process in which astrocytes play the role of medium carriers to provide adjustments for the involved synapses.

The interaction between several synapses, which is mainly mediated by the propagation of $\rm{Ca^{2+}}$ in cytosol within astrocytes, can be divided into three important phases. These phases will be concretely described in Section \uppercase\expandafter{\romannumeral3} so as to constitute the stigmergic system model. In this fundamental model, the strength of interaction between stigmergy agents is determined by the level of stimulations as well as the distance between them, which is consistent with the strength of interaction between synapses via astrocytes. Inspired by the morphological and functional changes in astrocytes during environmental enrichment, it is likely that the regulation of distance between stigmergy agents is critical to obtain stigmergy learning gain. Accordingly, the importance of the regulation is verified in two different stigmergic scenarios. 

 The remainder of this paper is organized as follows. In Section \uppercase\expandafter{\romannumeral2}, new discoveries on astrocytes in regulating synaptic transmission will be introduced and the existence of general stigmergy in the brain will be explored. In Section \uppercase\expandafter{\romannumeral3}, three important phases within the interaction between synapses will be described. The interaction is referred to set up the stigmergic system model and learning algorithm. In Section IV, the simulations on the performance of the proposed stigmergy learning are carried out in two different stigmergic scenarios, in order to verify its effectiveness and advantages. Finally, we conclude this paper with a summary.

\section{Stigmergy in the Brain}
In CNS, general stigmergy can be mapped into the interaction between synapses, which is mainly mediated by $\rm{Ca^{2+}}$ within astrocytes. As important medium carriers, astrocytes are coupled together by the gap-junction to comprise a nervous regulation.

\begin{figure*}[htbp]
	\begin{center}
		\includegraphics[width=0.8\textwidth]{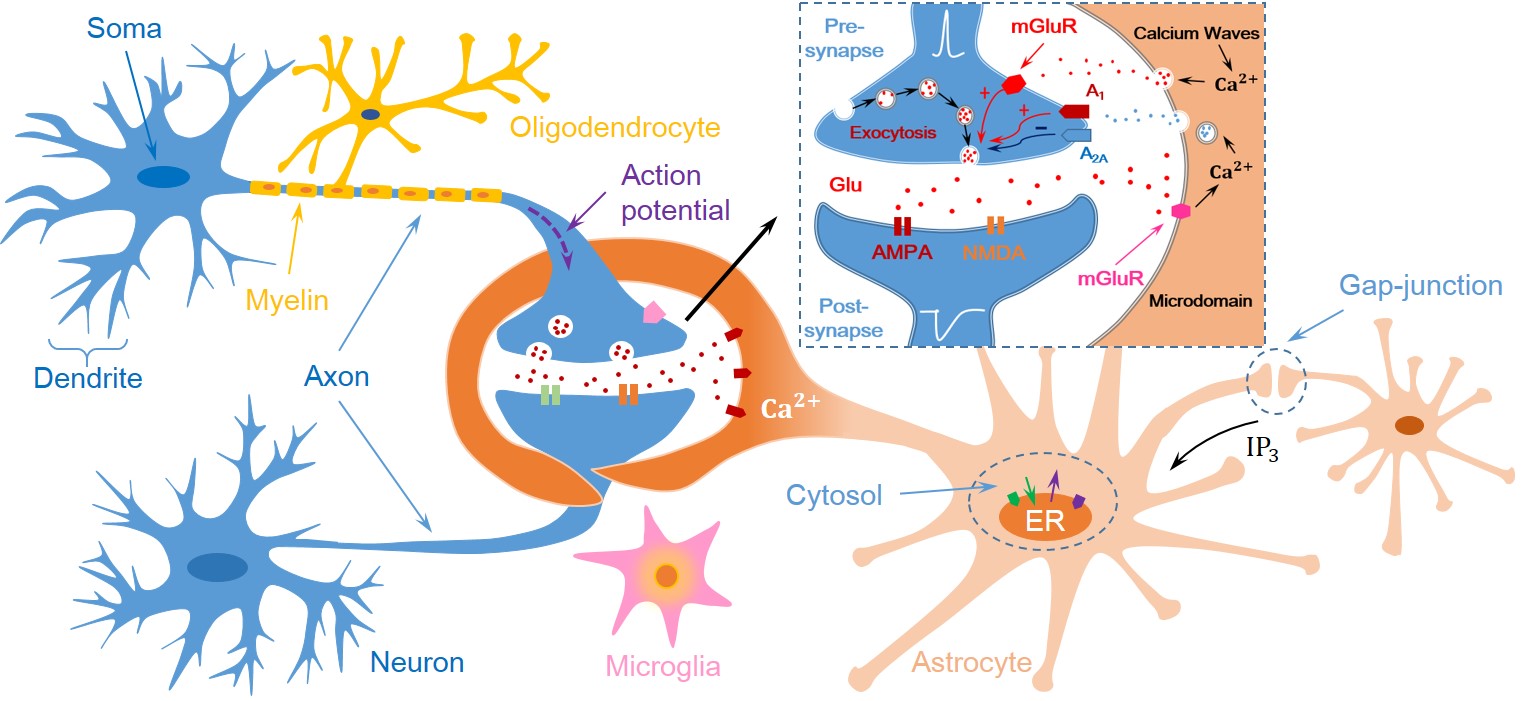}
		\caption{An intuitive diagram of the tripartite synapse.}
		\label{fig1}
	\end{center}
\end{figure*}

\subsection{Glial Cells in CNS}
In the process of nerve conduction, action potentials represented by the purple dotted arrow in Fig. 1 are conducted along the axon to the pre-synaptic terminal. Then a large quantity of neurotransmitters will be released into synaptic cleft through exocytosis. These molecules will diffuse and bind with various receptors on the surface of post-synaptic terminal. Besides, they can also diffuse and bind with receptors of surrounding glial cells, which will release neuromodulators in return\cite{Navarrete2011Basal}. In essential, there are three types of glial cells in CNS: microglia, oligodendrocytes, and astrocytes\cite{Correia2017On}. 

Microglia, as illustrated in Fig. 1, are macrophages in CNS. Their key roles are immune surveillance as well as responding to infections or other pathological states such as neurological diseases or injury\cite{Erny2016Communicating}\cite{Ransohoff2016How}. For the synaptic activity, microglia play the role of supervision and protection.

 Oligodendrocytes can contribute to the plasticity of nervous systems in the process of nerve conduction. An action potential needs to spend a certain amount of time reaching the pre-synaptic terminal. Many factors affect the conduction velocity, such as the thickness of myelin sheath, the axon diameter and the spacing and width of the Ranvier nodes\cite{Pajevic2013Role}. Increasing the thickness of myelin sheath can significantly improve the velocity, which helps to form the saltatory conduction. In this way, high-speed nerve pulses jump along the axon towards the pre-synaptic terminal, leading to a faster conduction. Oligodendrocytes play a critical role in this process because they can regulate the production of lecithin, which is an important substance for the compound of myelin\cite{Fields2015A}, as illustrated in Fig. 1. In this sense, the process of nerve conduction can be seen as the adjustment of the arrival time of different nerve pulses, which can be achieved by continuously changing the thickness of myelin sheath on each axon branch.

 Astrocytes are enriched with various receptors on the surface in order to support the implementation of different functions\cite{Sofroniew2010Astrocytes}. The phenomenon that the synaptic terminals as well as the cleft are wrapped by surrounding astrocytes gives rise to the structure of tripartite synapse\cite{Araque1999Tripartite}, which is illustrated with details in Fig. 1. In Fig. 1, the pre- and post-synaptic terminals are represented by the blue parts. The branch microdomain within astrocytes is represented by the yellow part. The transient calcium elevation in the microdomain may result from the binding with glutamate (Glu) which is released from the pre-synaptic terminal or the propagation of calcium waves from other microdomains. There is a large number of biochemical interactions between astrocytes and synapses, and various neuromodulators will be released from astrocytes due to the transient calcium elevation. These neuromodulators can act on purinergic $\rm{A_{2A}}$ (or $\rm{A_1}$) receptors on the pre-synaptic terminal to reduce (or increase) the number of exocytosis. Besides, $\rm{Ca^{2+}}$ with high concentration can diffuse to the other microdomains in the manner of calcium waves within astrocytes\cite{Bezzi2001A}. In this way, synapses wrapped by different branch microdomains can interact with each other while the propagation of calcium waves has constituted the main method of communications. 

\subsection{Regulation with Two Different Types}
When an action potential reaches the pre-synaptic terminal, a large quantity of Glu will be released into the synaptic cleft. These molecules will diffuse and act on metabotropic glutamate receptors (mGluRs) which are located at adjacent branch microdomains, evoking the production of a fix amount of $\rm{IP_3}$\cite{Nadkarni2004Dressed}. This process is schematically illustrated in the upper part of Fig. 2. The concentration of $\rm{IP_3}$ within astrocytes is believed as the key factor to evoke the elevation of intracellular calcium\cite{Mesiti2017Astrocyte}. Moreover, as shown in Fig. 2,  $\rm{IP_3}$ is considered as the second messenger to trigger the release of $\rm{Ca^{2+}}$ from endoplasmic reticulum (ER). ER can be considered as a reservoir with higher concentration of $\rm{Ca^{2+}}$ than that in cytosol. 

\begin{figure}[htbp]
	\begin{center}
		\includegraphics[width=0.4\textwidth]{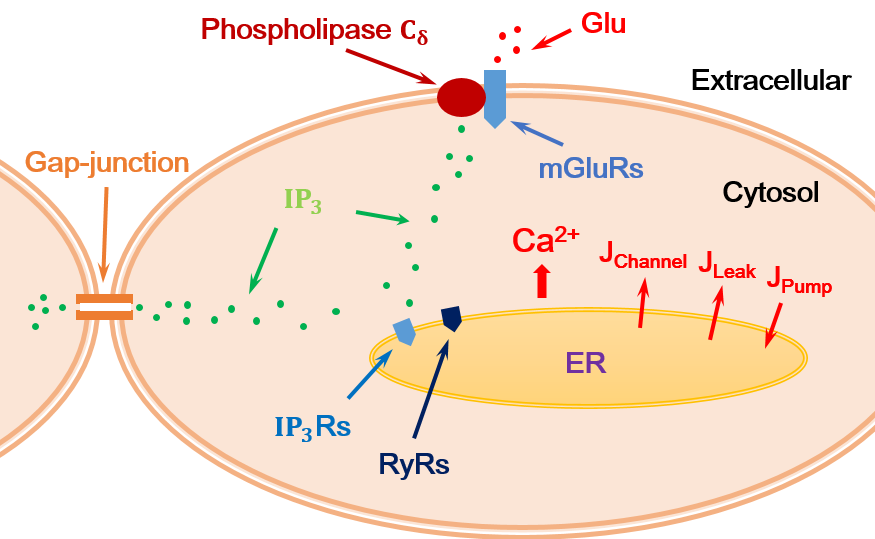}
		\setlength\abovecaptionskip{0pt}
		\setlength\belowcaptionskip{-5pt}
		\caption{The transient calcium elevation resulting from $\rm{IP_3}$.}
		\label{fig2}
	\end{center}
\end{figure}

A basic model in \cite{Li1994Equations} has been used to describe the dynamics of $\rm{Ca^{2+}}$ in cytosol due to the binding of $\rm{IP_3}$ with $\rm{IP_3}$ receptors ($\rm{IP_3Rs}$) in ER. There are three flows which are shown in the ER area in Fig. 2. $\rm{J_{Leak}}$ represents the leakage-flux of $\rm{Ca^{2+}}$ from ER into cytosol which is directly proportional to the concentration gradient of $\rm{Ca^{2+}}$ between ER and cytosol. $\rm{J_{Pump}}$ represents the pump-flux from cytosol into ER which needs to consume energy to maintain a concentration gradient. $\rm{J_{Channel}}$ represents the channel-flux from ER into cytosol which is generated due to the binding of $\rm{IP_3}$ with $\rm{IP_3Rs}$. The elevated concentration of $\rm{Ca^{2+}}$ in cytosol will further increase the open probability of $\rm{IP_3Rs}$ and ryanodine receptors (RyRs)\cite{Siekmann2015Data}, comprising of the mechanism known as Calcium-Induced Calcium-Release (CICR). Nevertheless, excessive concentration of $\rm{Ca^{2+}}$ in cytosol will bring down the open probability of $\rm{IP_3Rs}$ and RyRs, and the pump-flux $\rm{J_{Pump}}$ will become the main factor until a concentration gradient is re-established. 

Calcium waves can propagate between astrocytes to incur calcium oscillations\cite{Berridge1988Cytosolic}. There are many studies trying to describe and model the properties of the gap-junction between various astrocytes\cite{Goldberg2010Nonlinear}, as illustrated in Fig. 2. A large number of observations indicate that the gap-junction between astrocytes has a smaller conductance for $\rm{Ca^{2+}}$, but a larger one for $\rm{IP_3}$\cite{Pouilloux2006Anti}. Therefore, the above-mentioned $\rm{IP_3}$ is the main factor to promote the propagation of calcium waves between astrocytes. Besides, the activation of phospholipase $\rm{C_{\delta}}$ is also required for the propagation of calcium waves\cite{Pouilloux2006Anti}. An intuitive map of the transient calcium elevation resulting from $\rm{IP_3}$ in or between astrocytes is generalized in Fig. 2. 

On the other hand, neuroscience experiments have expressed different characteristics of $\rm{Ca^{2+}}$ between soma and microdomains. Typically, calcium elevations occurring in the microdomains are much more frequent and transient than those in the soma\cite{Kanemaru2014In}. Researchers in\cite{Shigetomi2012TRPA1} indicated that there should be Transient Receptor Potential Ankyrin type 1 (TRPA1) or receptor-gated $\rm{Ca^{2+}}$-permeable ions channels in the astrocyte membrane, through which $\rm{Ca^{2+}}$ could flux into the cell from the extracellular matrix. Recent studies indicate that there are actually two different types of regulations within astrocytes\cite{Bazargani2016Astrocyte}. The short-range regulation in response to low-intensity stimulus is induced by rapid and short-term calcium elevations. The long-range regulation in response to high-intensity stimulus is induced by slow and long-term calcium elevations. The former provides the regulation within the scale of several synapses locally, and $\rm{Ca^{2+}}$ influx through the receptor-gated ions channels can be the main factor. The latter provides the regulation among different astrocytes with the propagation of calcium waves through the gap-junction, and $\rm{IP_3}$ can be the main factor. In this paper, we focus our attention on the short-range regulation. 

\begin{table}
	\begin{center}
		\caption{The Main Symbols and Acronyms.}
		\setlength\abovecaptionskip{-5pt}
		\setlength\belowcaptionskip{-5pt}
		\label{tb2}
		\begin{tabular}{c||c||}
			\hline
			Acronym         & Description     \\	
			\hline
			CNS   &    Central Nervous System\\	
			Glu   &   Glutamate  \\
			$\rm{IP_3}$    & Inositol-1,4,5-triphospate  \\
			ER     & Endoplasmic Reticulum \\
			$\rm{IP_3Rs}$ & Inositol-1,4,5-triphospate receptors \\
			RyRs  &  Ryanodine receptors \\
			CICR  & Calcium Induced Calcium Release   \\ 
			mGluRs  & Metabotropic glutamate receptors  \\
			AMPA & $\alpha$-Amino-3-hydroxy-5methy1-4-isoxazolepropionic acid \\
			NMDA &  N-methil-D-aspartic acid \\
			TRPA1 & Transient Receptor Potential Ankyrin type 1 \\
			SOM   &  Self-Organizing Mapping \\
				\hline
		\end{tabular}
	\end{center}
\end{table}

\subsection{Astrocytes as Regulation Networks}
Astrocytes occupy a fundamental position in the synaptic activity. It is suggested that the efficiency of synaptic transmission through the pre-synaptic terminal will be greatly decreased without the calcium signal\cite{Navarrete2011Basal}. The microdomain with elevated calcium will generate an effect for the wrapped synapse. Many researchers tried to decode the calcium signal\cite{De2009Multimodal}\cite{Nadkarni2004Dressed}. Receptors on the membrane of post-synaptic terminal have low affinity. But the interaction between synapses and astrocytes is granted by receptors with high affinity and slow desensitization. It means that the influence from synapses and astrocytes will not disappear immediately.

In general, the arriving time of consequent action potentials at a certain synapse can be regarded as a discrete-time pulse sequence. Each of them can change the synaptic state into excitatory or inhibitory. The synaptic state change will generate calcium elevations in surrounding microdomains which will provide feedback in return. The synapse will gradually recover to its original state until the arrival of the next action potential. In this situation, the duration of the elevation depends on the length of the arriving time interval, and a shorter one will produce a longer duration. Therefore, the level of synaptic activities can be measured by the level of calcium elevations. Researchers in \cite{Araque2014Gliotransmitters} found that increasing the level of synaptic activities would lead $\rm{Ca^{2+}}$ diffusing into the adjacent microdomains, and a persistent high level would eventually make $\rm{Ca^{2+}}$ be full of the whole astrocyte, which is depicted in Fig. 3. In Fig. 3, (a) represents the response of astrocytes under low intensity stimulus. The red solid arrow represents the diffusion direction of $\rm{Ca^{2+}}$ while the black dotted arrow represents the feedback effect. Fig. 3 (b) is the intermediate result of increasing the intensity of stimulus. Fig. 3 (c) shows the final diffusion effect caused from a synapse which is stimulated by consequent action potentials.

\begin{figure}[!h]
	\begin{center}
		\includegraphics[width=0.48\textwidth]{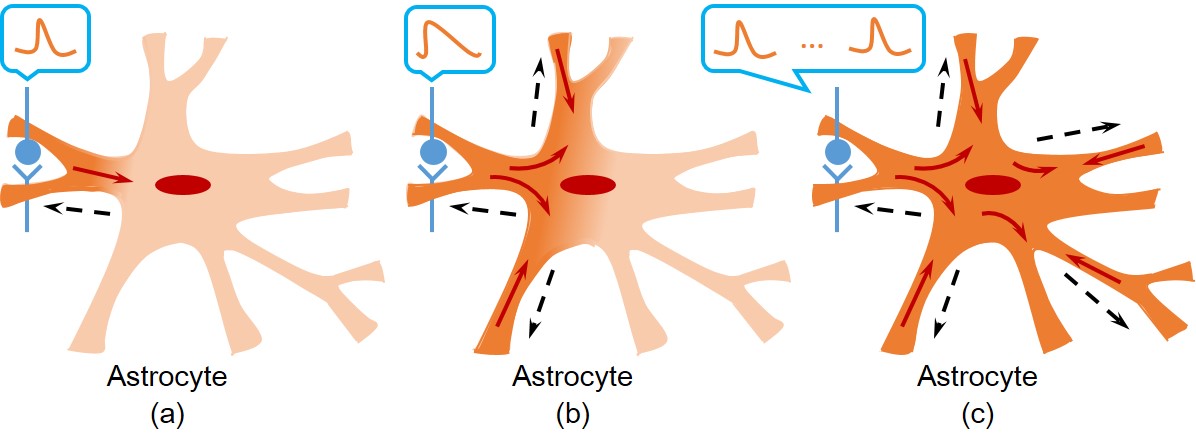}
		\setlength\abovecaptionskip{0pt}
		\setlength\belowcaptionskip{-5pt}
		\caption{Different levels of calcium elevations caused by different levels of synaptic activities.}
		\label{fig3}
	\end{center}
\end{figure}

Many low levels of calcium elevations, which can generate the above-mentioned short-range regulations within the scale of several synapses, can jointly comprise a large state change which can be detected in the whole astrocyte. Furthermore, these calcium elevations form  a $\rm{Ca^{2+}}$ concentration map within astrocytes. This consideration brings about the concept that astrocytes can act as an encoder to encode the temporal properties of synaptic activities into spatial patterns. Meanwhile, astrocytes can be regarded as a spatial regulation network, in which the activity of a synapse can influence the states of other synapses not only in the adjacent area, but also in the distant regions with the help of calcium waves. As described in Fig. 4, this regulation network can provide a cross regulation for nervous system. Different from the nerve conduction, synapses by means of the spatial regulation network of astrocytes can activate other neighbor neurons between which there is no direct connection.  

\begin{figure}[!h]
	\begin{center}
		\includegraphics[width=0.3\textwidth]{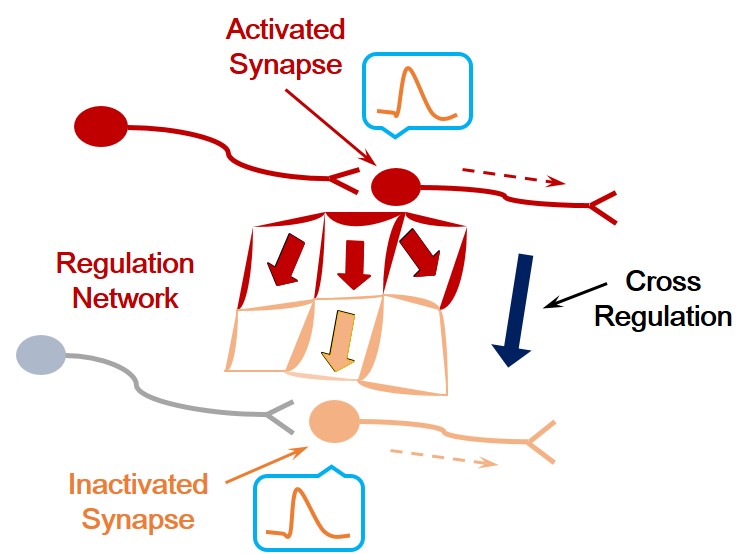}
		\setlength\abovecaptionskip{0pt}
		\setlength\belowcaptionskip{-5pt}
		\caption{A cross regulation provided by astrocytes for nervous system.}
		\label{fig4}
	\end{center}
\end{figure}

\subsection{Stigmergy in the Brain}
In the hippocampal stratum radiatum, the detailed 3D reconstruction work shows that 80\% synapses are coupled with the branch microdomains, and astrocytes almost completely wrap synapses which are rich in docked vesicles\cite{Ventura1999Three}. A large number of synapses with certain functions are coupled together through astrocytes to form a potentially collaborative nervous system. Calcium waves comprise the main method of communications between synapses which are not directly connected. Accordingly, general stigmergy can be mapped into the mechanism of cooperative interaction between synapses. 

In the brain's nervous system, various synapses can be regarded as different stigmergy agents, and a map of $\rm{Ca^{2+}}$ concentration within astrocytes can be regarded as the medium. Action potentials can change the synaptic state into excitatory or inhibitory, which will generate different levels of calcium elevations in the corresponding microdomains. This process can be regarded as leaving traces in the medium as in general stigmergy. With the help of $\rm{Ca^{2+}}$ and $\rm{IP_3}$, calcium waves can expand its influence throughout astrocytes. The superposition of different calcium elevations is linear, thus the effect of local traces can be integrated to adjust the whole stigmergic environment. An illustrative comparison between general stigmergy and the mechanism of stigmergic interactions between synapses and astrocytes is illustrated in Fig. 5.

\begin{figure}[!h]
	\begin{center}
		\includegraphics[width=0.47\textwidth]{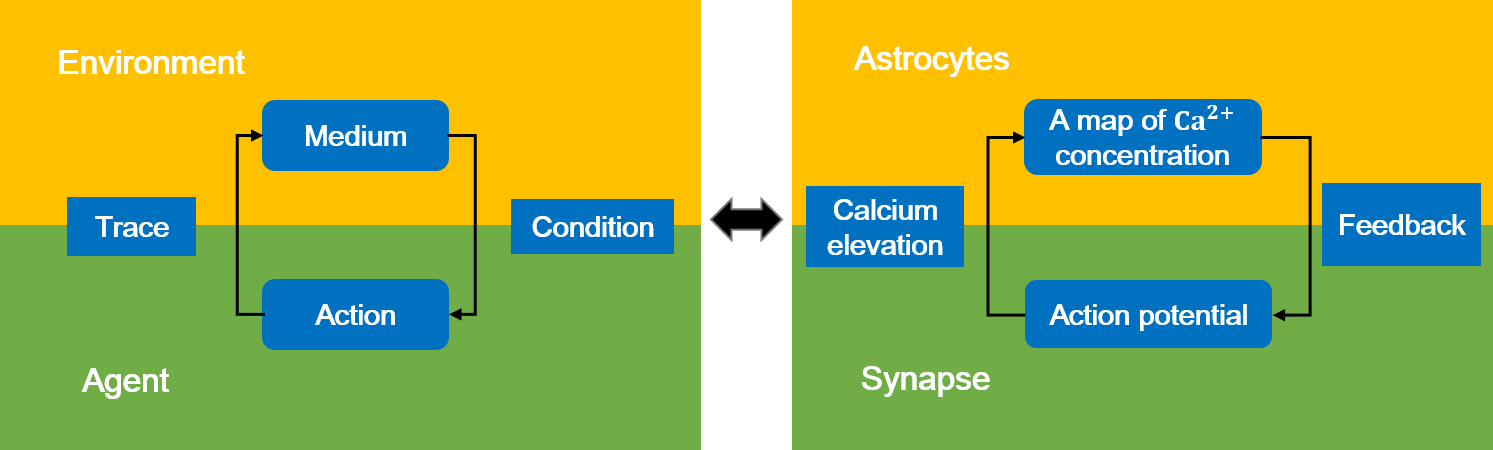}
		\setlength\abovecaptionskip{0pt}
		\setlength\belowcaptionskip{-5pt}
		\caption{A comparison between general stigmergy and the mechanism of stigmergic interactions between synapses and astrocytes.}
		\label{fig5}
	\end{center}
\end{figure}

Astrocytes can be regarded as significant medium carriers, which maintain the map of $\rm{Ca^{2+}}$ concentration. Astrocytes can also provide the regulation for the involved synapses, whose implementation benefits from a large number of receptors with different types between synapses and astrocytes. This effect can be regarded as the condition provided by the medium for stigmergy agents. Besides, the concentration of $\rm{Ca^{2+}}$ in astrocytes will decay with time, which comprises a negative feedback loop and provides stability for the nervous system with controlled cycles. Because of a limited range of influence, only regulations reflecting the right condition of the nervous system will superpose and have a longer duration. Through this kind of stigmergic process, astrocytes integrate the calcium elevations generated by different synapses and provide cross-regulation for various individual synapses in the nervous system.

\section{Stigmergy Learning Mechanism and Model}
Based on the aforementioned analyses, the interaction process within the scale of several synapses, which are not directly connected, can be modeled. The implementation of their interactions, which is regarded as the short-range regulation, mainly relies on the propagation of $\rm{Ca^{2+}}$  throughout astrocytes. Hereinafter, we divide this process into three phases which are described in Fig. 6. 

The first phase which represents the generation of calcium elevation resulting from the activated synapse in the microdomain is indicated by \uppercase\expandafter{\romannumeral1} in Fig. 6. At first, the release of Glu due to the arriving of an action potential is modeled by\cite{Destexhe1994Synthesis}:

\begin{equation}
[T_{Neur}]=\frac{T_{max}}{1+\exp (-\frac{V_d-V_{base}}{K_N})}
\end{equation}
where $[T_{Neur}]$ is the concentration of Glu in synaptic cleft, and $T_{max}$ represents its maximum. $V_d$ is the voltage of dendrite in the Pinsky-Rinzel model\cite{Pinsky1995Intrinsic}.  $V_{base}$ and $K_N$ are parameters used to modify the sigmoid function curve. Then Glu will diffuse and act on receptors on the membrane of branch microdomain to increase the concentration of $\rm{Ca^{2+}}$:

\begin{equation}
\frac{d[Ca^{2+}]}{dt}=\frac{v_{Ca}*[T_{Neur}]^n}{k_{Ca}^n+[T_{Neur}]^n}-\frac{1}{\tau _{Ca}}([Ca^{2+}]-[Ca^{2+}]^*)
\end{equation}
where $v_{Ca}$ and $k_{Ca}$ are regulating parameters. $n$ is an adjusting factor. $\tau _{Ca}$ is a decay constant. $[Ca^{2+}]^*$ represents the concentration of $\rm{Ca^{2+}}$ at equilibrium in cytosol. The first item in the equation expresses the increment of $\rm{Ca^{2+}}$ concentration in the microdomain. The second item indicates that the concentration also decreases with time because of the concentration gradient of $\rm{Ca^{2+}}$ between cytosol and the extracellular matrix.

\begin{figure}
	\begin{center}
		\includegraphics[width=0.4\textwidth]{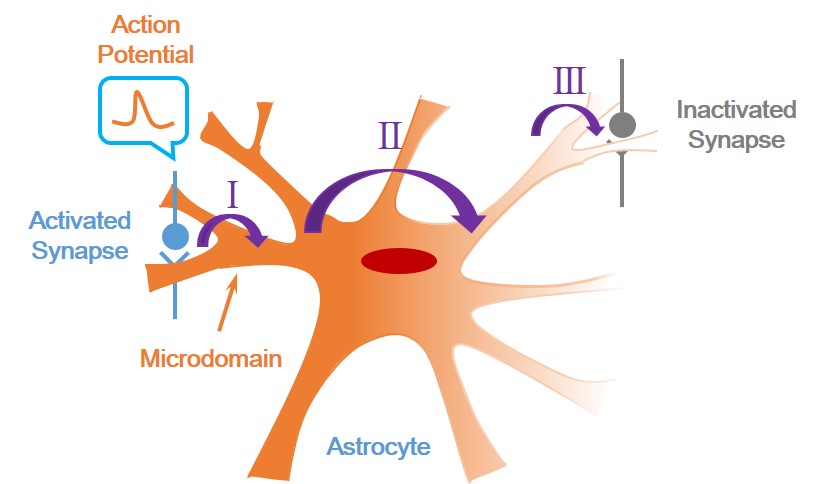}
		\setlength\abovecaptionskip{0pt}
		\setlength\belowcaptionskip{-5pt}
		\caption{Three phases included in the interaction between synapses. They are respectively numbered by \uppercase\expandafter{\romannumeral1}, \uppercase\expandafter{\romannumeral2} and \uppercase\expandafter{\romannumeral3}.}
		\label{fig6}
	\end{center}
\end{figure}

The second phase which considers the passive diffusion of $\rm{Ca^{2+}}$ from one microdomain to others is indicated by \uppercase\expandafter{\romannumeral2} in Fig. 6. The passive diffusion of $\rm{Ca^{2+}}$ can be calculated by the Telegraph Equation\cite{Ali2005Relativistic}:

\begin{equation}
\tau_d\frac{\partial^2c(x,t)}{\partial t^2}+\frac{\partial c(x,t)}{\partial t}=D\nabla^2c(x,t)+b(x_0,t)
\end{equation}
where $\tau_d$ is the relaxation factor accounting for a finite propagation speed. $c(x,t)$ is the concentration of $\rm{Ca^{2+}}$ at location $x$ and time $t$. $D$ is the diffusion coefficient. Furthermore, $b(x_0,t)$ representing the change rate of concentration at the initial point is given by\cite{Pierobon2010A}:

\begin{equation}
b(x_0,t)=\frac{dc(x_0,t)}{dt}
\end{equation}

The third phase which considers the regulation provided by astrocytes with elevated calcium for synapses is indicated by \uppercase\expandafter{\romannumeral3} in Fig. 6. The relationship between the concentration of $\rm{Ca^{2+}}$ in the corresponding microdomain and the amplitude of slow inward currents in the pre-synaptic terminal has been used to describe the regulation\cite{Nadkarni2004Dressed}:

\begin{equation}
	\begin{aligned}
		I_{current}&=k_I\Theta(\ln y)\ln y\\
		y&=[Ca^{2+}]-I_{th}
	\end{aligned}
\end{equation}
With regard to Eq. (5), there is a threshold value $I_{th}$ for the concentration of $\rm{Ca^{2+}}$ before providing a regulation for the pre-synaptic terminal. $k_I$ is a scale factor. $\Theta$ represents the Heaviside function. A natural logarithmic function is used to describe the strength of regulation, which will generate an effect until the concentration of $\rm{Ca^{2+}}$ is below a certain threshold. Therefore, this function determines a scope for the propagation of calcium waves.

\begin{figure}[!h]
	\begin{center}
		\includegraphics[width=0.5\textwidth]{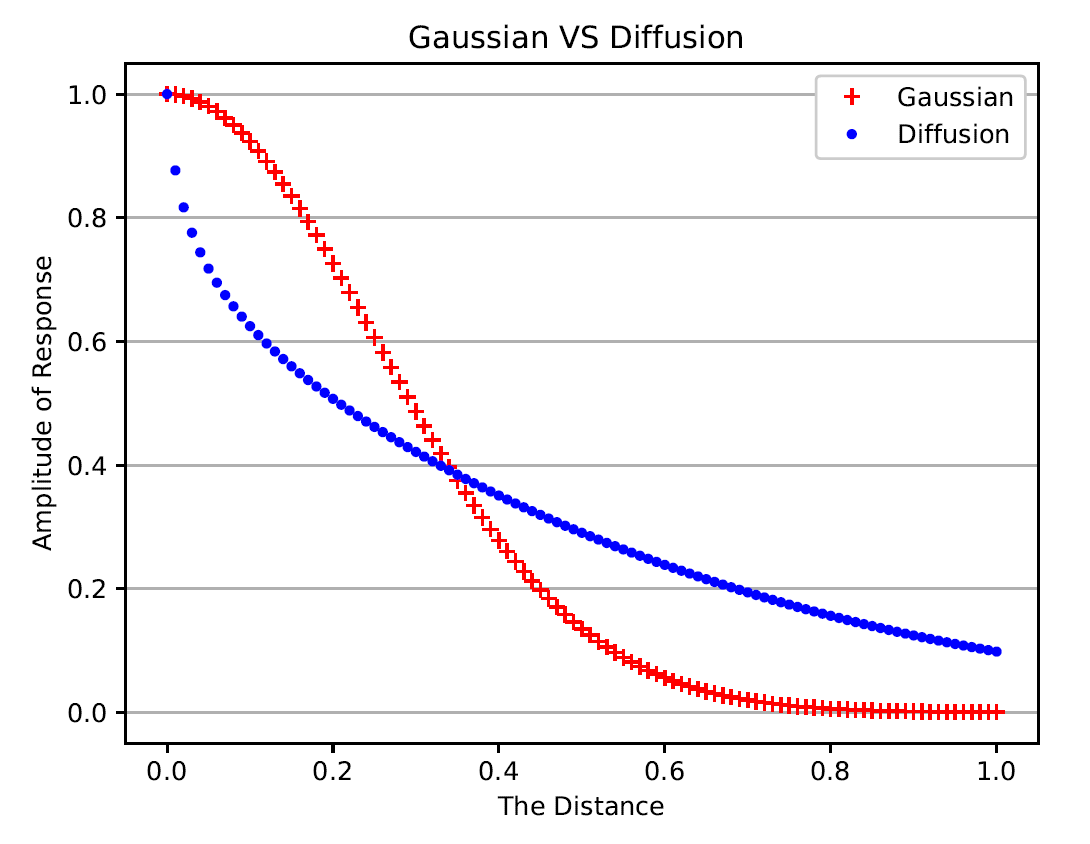}
		\setlength\abovecaptionskip{0pt}
		\setlength\belowcaptionskip{-5pt}
		\caption{The strength of synaptic interaction at different distances.}
		\label{fig7}
	\end{center}
\end{figure}

We can further integrate the above three phases together to describe the strength of synaptic interaction at different distances, which is represented by the Diffusion curve in Fig. 7.
In Fig. 7, the Diffusion curve has been normalized to match the degree of the Gaussian function. Compared with the Gaussian function which is widely used as neighborhood function in Self-Organizing Mapping (SOM) of neural networks, the Diffusion curve has a similar downward trend. Regardless of the initial stimulus intensity, synapses with larger distances will have smaller amplitude of responses through the Diffusion process. We will take advantage of this relationship between synapses to coordinate the behaviours of stigmergy agents. 

More specifically, rooted in the above three interactive phases, a stigmergic learning mechanism is proposed and illustrated in Fig. 8, in which the communications between different stigmergy agents (i.e. synapses) represented by different colors are indirect. When getting a stimulus input, a stigmergy agent will leave traces (i.e. calcium elevation) which is expressed by the red solid arrow in the outside environmental medium to affect the state of other agents. As illustrated in Fig. 8, the amplitude of response for the interactive influence indicated by the blue dotted arrow is determined by the inter-synapse distance $x$ between stigmergy agents as well as the intensity of initial stimulus $s$. In the nervous system, the intensity of initial stimulus is consistent with the level of synaptic activity while the synaptic distance is determined by the distance between the coupled branch microdomains within astrocytes. 

Environmental enrichment, which is the stimulation of the brain by its physical and social surroundings, is known to induce the increases in synaptic and spine densities. Researchers in \cite{Viola2009Morphological} found that great changes would present in astrocytic morphology and a large number of branch microdomains would appear in this process. Astrocytes by virtue of these emerging microdomains can coordinate synaptic activities and change the amplitude of responses between these neurons. Therefore, the inter-synapse distance adaptation between stigmergy agents in the stigmergic learning mechanism can be leveraged and well-regulated to support mutual efficient collaboration.

\begin{figure}
	\begin{center}
		\includegraphics[width=0.4\textwidth]{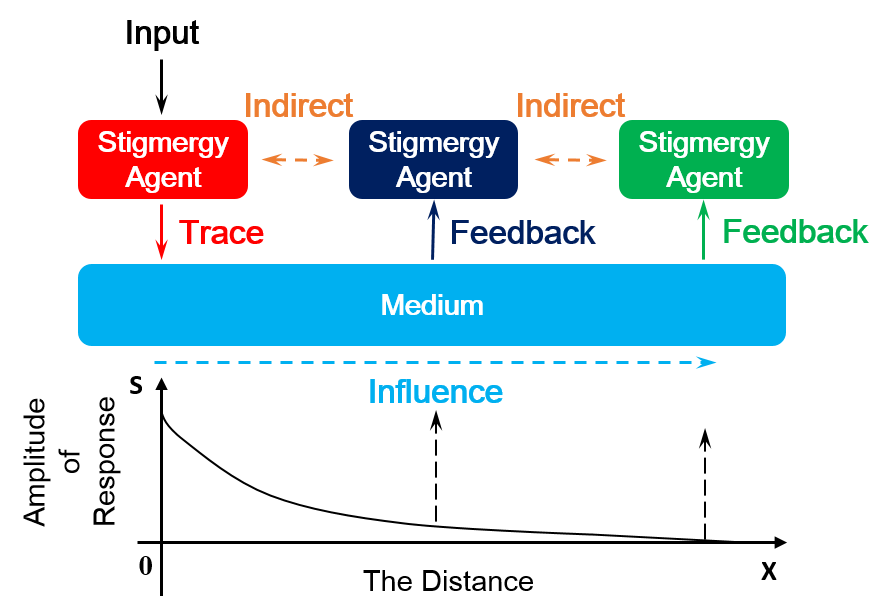}
		\setlength\abovecaptionskip{0pt}
		\setlength\belowcaptionskip{-5pt}
		\caption{The stigmergic learning mechanism.}
		\label{fig8}
	\end{center}
\end{figure}

Furthermore, a multi-agent cooperation approach is considered to take advantage of the regulation of inter-synapse distance between stigmergy agents to obtain stigmergic learning gain. Within this approach, stigmergy agents are continuously selected out for certain tasks in each turn until their common object requirements are reached. In particular, we leverage and modify the strategy proposed in \cite{Dorigo2000Ant} in each selection round:

\begin{equation}
p_{i,j}(t)=\frac{s_j^n(t)}{s_j^n(t)+ \alpha \theta_{i,j}^n(t)+\beta \varphi_{i,j}^n(t)}
\end{equation}
\begin{equation}
p_{i,j}(t)=\frac{s_j^n(t)}{s_j^n(t)+ \alpha \theta_{i,j}^n(t)*\beta \varphi_{i,j}^n(t)}
\end{equation}
where $p_{i,j}(t)$ is the probability of $i_{th}$ stigmergy agent being selected for $j_{th}$ task. $s_j(t)$ is the emergency degree of $j_{th}$ task. $\alpha$, $\beta$ and $n$ are adjusting factors. $\theta_{i,j}(t)$ is the state value of $i_{th}$ agent for $j_{th}$ task. $\varphi_{i,j}(t)$ is a heuristic factor. Comparing with Eq. (6), in the modified strategy Eq. (7), ``$+$" has been changed into ``$\ast$" in order to remove the asynchronous variation which will cause the jitter at steady state. After each action, the update process of $s_j(t)$ is modified as:

\begin{equation}
R_j(t)=R_j(t-1)+\sum_{m\in S_j(t-1)}r_{m,j}(t)
\end{equation}
\begin{equation}
s_j(t)=R_j(t)/T_j
\end{equation}
where $r_{m,j}(t)$ is the reward that $m_{th}$ agent obtains in $j_{th}$ task at time $t$ from the outside environmental medium. $R_j(t)$ is the sum of all rewards at time $t$. $T_j$ is the expected object requirement for task $j$. $S_j(t-1)$ is the set of stigmergy agents which participate in $j_{th}$ task at time $t-1$. As more stigmergy agents participate in this task, $s_j(t)$ will get bigger (approach to 1) and thus provide a stimulated collaboration with higher intensity. 

The state value of different stigmergy agents for the same task can be different in Eq. (7). After taking an action, this state value will be updated according to the following equations:

\begin{equation}
\theta_{i,j}(t)=\theta_{i,j}(t-1)+\Delta \theta_{i,j}(t-1)
\end{equation}
\begin{equation}
\Delta \theta_{i,j}(t-1)=\rho_1*(\frac{1}{|S_j(t-1)|}\sum_{m\in S_j(t-1)}r_{m,j}-r_{i,j})
\end{equation}
where $\rho_1$ is a scale factor. $\Delta \theta_{i,j}(t-1)$ can be positive or negative, which corresponds to a low or high reward respectively. According to the proposed stigmergic learning mechanism, the state change of a stigmergy agent will expand its influence to affect the state of other agents. Therefore, the state value should be further updated by:

\begin{equation}
\theta_{i,j}(t) = \theta_{i,j}(t) +\overline{\Delta \theta_{i,j}(t)}
\end{equation}
\begin{equation}
\overline{\Delta \theta_{i,j}(t)}=\sum_{k\in \pi_i(t-1)}D(d_{k,i}(t-1))*\Delta \theta_{k,j}(t-1)*\rho_2
\end{equation}
where $\rho_2$ is a scale factor. $\pi_i(t-1)=\{X_k|k\ne i,d_{k,i}(t-1)<d_{th}\}$. $d_{k,i}(t-1)$ is the inter-synapse distance between $k_{th}$ and $i_{th}$ agent at time $t-1$. $d_{th}$ is a threshold value for the inter-synapse distance. $D(\cdot)$ represents the interaction process which includes the above-mentioned three phases. Here we use $\Delta \theta_{k,j}(t-1)$ to represent the intensity of stimulus provided by $k_{th}$ agent for $i_{th}$ agent, which will be discounted by their synaptic distance. The distance between stigmergy agents, which describes the strength of synaptic interaction, can be regulated according to the feedback. Accordingly, we further put forward the following scheme to regulate the distance after each action:

\begin{equation}
d_{k,i}(t) = \left\{
\begin{array}{lcl}
{d_{k,i}(t-1) - factor}, &\text{if} \   \phi >0 \\
{d_{k,i}(t-1) + factor}, &\text{otherwise}   
\end{array}  
\right.
\end {equation}
\begin{equation}
	\phi = \Delta \theta_{i,j}(t)*\Delta \theta_{k,j}(t-1)
\end{equation}
where $factor$ is a constant. To some extent, the distance between stigmergy agents also represents the similarity of these agents participating in the same task. Therefore, the strength of interactions will be larger if the similarity between two stigmergy agents is higher. The regulation of inter-synapse distance can adjust the strength of synaptic interactions  between stigmergy agents and thus bring the systematic stigmergy learning gain.

\section{Numerical Simulation and Results}
In order to verify the effectiveness and advantages of the proposed stigmergy learning model, a number of numerical simulations with different kinds of tasks have been carried  out. 

\subsection{The Stigmergy Learning Gain}

In the first simulation, there is only one task to be finished. During the initialization, the distance between various neural agents is set as the median value in the range, which can represent the similarity of these agents participating in this task. The same method is also applied to the setting of the state value. A random reward is assigned to each neural agent which will not be changed during the whole simulation process. Several neural agents are allowed to take an action together as a batch. There is a fixed cost for each action which is the same for all agent individuals. Besides, an ability value is randomly assigned to each neural agent indicating the number of actions it can still take, which is also utilized as the heuristic factor in Eq. (7) and normalized to match the degree of the state value.

 In each turn, several neural agents are selected out according to the selection probability (Eq. (7)) as a batch to participate in the target task until the overall reward is above the object requirement. After each turn, a feedback is returned, which equals to the sum of all rewards of neural agents that are selected out in last turn. This feedback is used to regulate the distance between neural agents. The target of the simulation is to satisfy the object requirement with the maximum utility value (reward/cost). The main parameters within the simulation are shown in Table \uppercase\expandafter{\romannumeral2}.

\begin{table}[!hbp]
	\begin{center}
		\caption{The Main Parameters.}
		\setlength\abovecaptionskip{-0pt}
		\setlength\belowcaptionskip{-0pt}
		\label{tb1}
		\begin{tabular}{c|c}
			\toprule
			Item         & Description   \\
			\hline		
			$\rm{Agent\_number}$     & $30$  \\
			\hline
			$\rm {Object\_requirement}$     & $1100$ \\
			\hline
			$\rm{Batch\_size}$  & $5$   \\ 
			\hline
			$\rm{Agent\_reward}$  & $[1,10]$   \\
			\hline
			$\rm{Agent\_ability}$  & $[50,120]$    \\	
			\hline
			$\rm{Cost}$ & $10$ \\
			\hline
			$\rm{\alpha}$ & $2$\\
			\hline
			$\rm{\beta}$ & $2$\\
			\hline
			$\rm{n}$ & $2$\\
			\hline
			$\rm{\rho_1}$ & $0.001$\\
			\hline
			$\rm{\rho_2}$ & $1$\\
			\hline
			$\rm{factor}$ & $0.5$\\
			\bottomrule
		\end{tabular}
	\end{center}
\end{table}

According to the selection probability, neural agents with higher rewards are assumed to have smaller state values and thus more likely to be selected. These neural agents will shorten the distance between those agents with the same higher rewards to form a cluster. Based on the proposed stigmergy learning model (Eq. (7) - (15)), aiming at forming a spatial neural cluster, the task is repeatedly carried out for 500 times to regulate the distance between neural agents. The state value and the average distance of neural agents are given in Fig. 9. In Fig. 9, neural agents are arranged in descending order according to the state values. Similarly, the average distance which represents the average distance value of a neural agent from all the other agents is arranged according to the corresponding order. It can be observed that the neural agents  with lower state values will have smaller average distances. Therefore, a neural cluster can be automatically formed by agents with higher rewards. Members in the neural cluster have smaller distances than the others, whose stimulus resulting from the state change can thus be more easily responded.

\begin{figure}[!h]
	\begin{center}
		\includegraphics[width=0.45\textwidth]{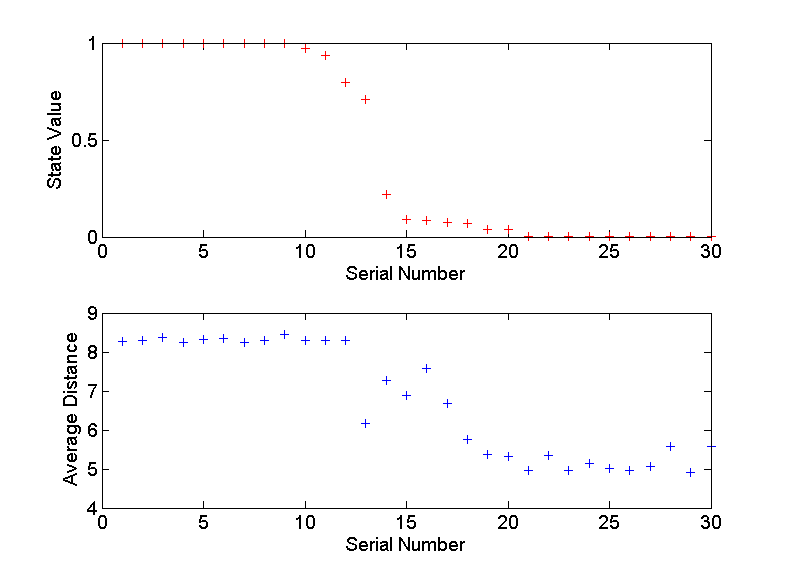}
		\setlength\abovecaptionskip{0pt}
		\setlength\belowcaptionskip{-0pt}
		\caption{The state value and the average distance of neural agents.}
		\label{fig9}
	\end{center}
\end{figure}

\begin{figure}
	\begin{center}
		\includegraphics[width=0.45\textwidth]{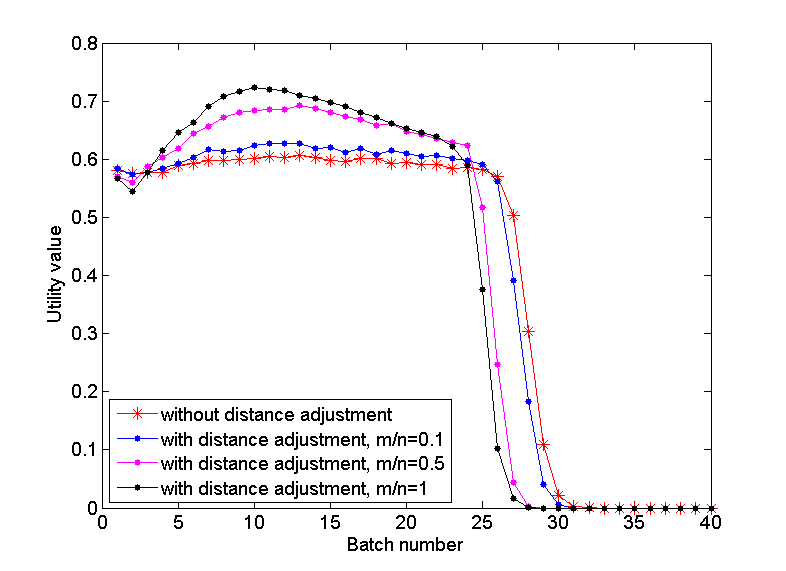}
		\setlength\abovecaptionskip{0pt}
		\setlength\belowcaptionskip{-0pt}
		\caption{The system gain provided by the regulated distance between neural agents.}
		\label{fig10}
	\end{center}
\end{figure}

The obtained neural cluster can be used to generate the stigmergy learning gain. We have compared the utility value of each batch with that in general stigmergy, in which the distance adjustment between stigmergic agents as well as the formation of the neural cluster are not taken into account. The comparison results are provided in Fig. 10. In Fig. 10, the curve without distance adjustment represents the utility value of the traditional stigmergy mechanism while the curves with distance adjustment represent the utility values of the proposed stigmergy learning model. $m$ in Fig. 10 represents the maximum of $\overline{\Delta \theta_{i,j}(t)}$ while $n$ represents the maximum of $\Delta \theta_{i,j}(t)$. Because of the limitation of the ability value, the last few rounds in each scheme have to adopt neural agents with lower rewards, which cause the decline of each curve in the end. With the existence of the regulated distance, the task is started  with higher efficiency and completed earlier. Accordingly, neural agents with higher rewards are more easily selected in the task and further activate those with the same higher rewards because of the neural clustering merit. Therefore, as a key in the proposed stigmergy learning model, the regulation of distance between various neural agents through spatial clustering can bring expected gain for the learning system.

\subsection{The Impact of Distance Regulation}
Different from the first simulation, the aim of the second simulation is to test if we can adjust current states of neural agents to converge to the target pattern (e.g. a target picture of Arabic numerals). The selection of neural agents is different with that in the first simulation, namely, $120$ agents are randomly selected out to form a neural group in each turn. The current state of each neural agent is redefined by:

\begin{equation}
	y_j=\Theta(\sum_{j\ne i}v_i*D(d_{i,j})-base)
\end{equation}
where $base$ is a constant. $\Theta$ represents the Heaviside function. $v_i$ represents the input of $i_{th}$ agent.  $y_j$ represents the output of $j_{th}$ agent. $y_{j}=1$ means that the current state of $j_{th}$ agent is excitatory while $y_{j}=0$ means that the current state of $j_{th}$ agent is inhibitory. Self-feedback is not considered here. The relationship between neural agents is illustrated in Fig. 11 (a), in which distances between neural agents are directed. Regardless of the initial input, according to Eq. (16), the current state of a neural agent is determined by the distance from the other agents.

\begin{figure}[!h]
	\begin{center}
		\includegraphics[width=0.45\textwidth]{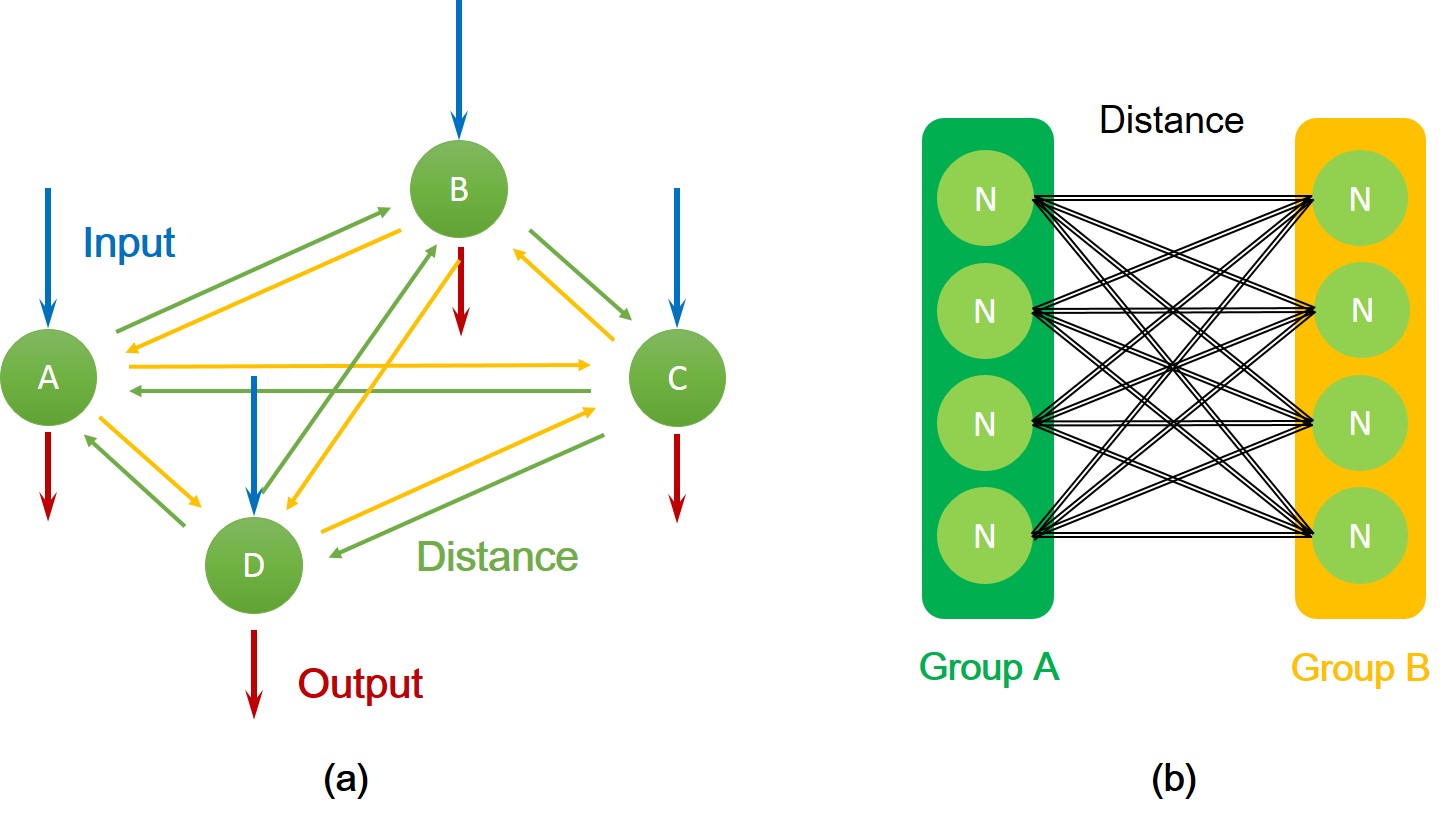}
		\setlength\abovecaptionskip{0pt}
		\setlength\belowcaptionskip{-0pt}
		\caption{The relationship between neural agents.}
		\label{fig11}
	\end{center}
\end{figure}

In each turn, all neural agents will be given a unit input to determine the current states of neural agents in the group. These states can be further used to calculate the feedback which equals to the sum of all rewards of neural agents in the group. The reward for each state of neural agent is provided in Table \uppercase\expandafter{\romannumeral3}. In Table \uppercase\expandafter{\romannumeral3}, the value of a pixel refers to the binary value in the original picture. The size of the original picture is $28\times28$, in which each pixel is represented by the state of a neural agent in the corresponding location. 

\begin{table}[!h]
	\begin{center}
		\caption{The Reward For Each State.}
		\setlength\abovecaptionskip{-0pt}
		\setlength\belowcaptionskip{-0pt}
		\label{tb2}
		\begin{tabular}{c|c|c}
			\toprule
			state         & pixel    & reward \\
			\hline		
			1     & 1  & 0\\
			\hline
			1     & 0  & -1\\
			\hline
			0  & 1    &+1\\ 
			\hline
			0  & 0   & 0\\
			\bottomrule
		\end{tabular}
	\end{center}
\end{table}

According to the selection method, different neural groups that may contain the same members are selected out in different turns. Neural agents in each group will regulate the distance to adjust their current states according to the corresponding feedback. Concretely, as shown in Fig. 11 (b), we compare the feedback of two neural groups in two continuous turns and change the distance between any two members in different neural groups according to the result. The distance from neural agents in the group with larger feedback to the others in the group with smaller one will increase a constant in each turn. As mentioned before, the distance between neural agents describes the strength of interactions between them and a shorter distance indicates a higher strength. In this way, neural agents which should be excitatory for the target pattern will get shorter distances and be more easily activated. 

\begin{figure}[!h]
	\begin{center}
		\includegraphics[width=0.5\textwidth]{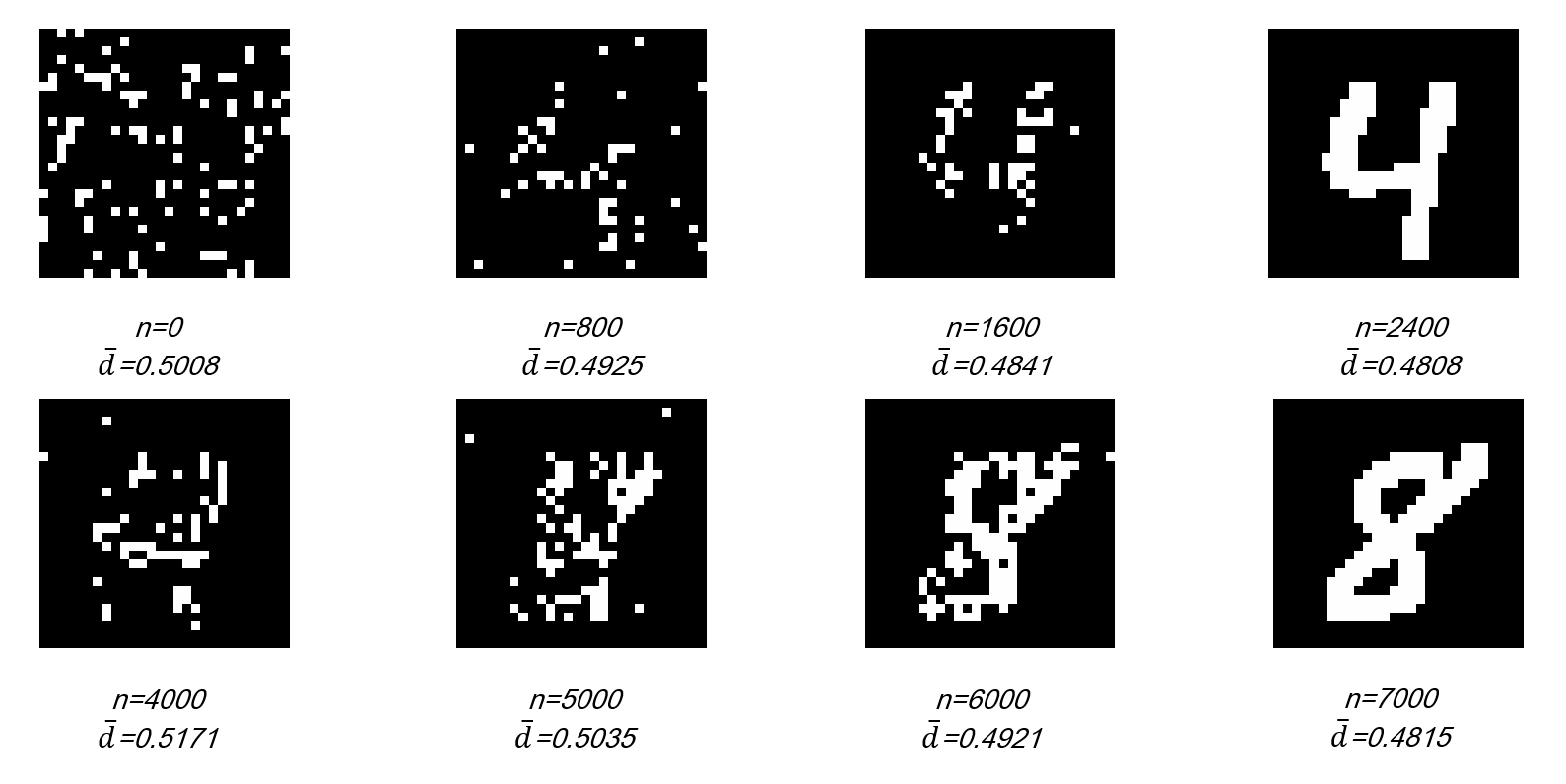}
		\setlength\abovecaptionskip{0pt}
		\setlength\belowcaptionskip{-0pt}
		\caption{The learning process of the sitgmergy learning system.}
		\label{fig12}
	\end{center}
\end{figure}

\begin{figure}
	\begin{center}
		\includegraphics[width=0.5\textwidth]{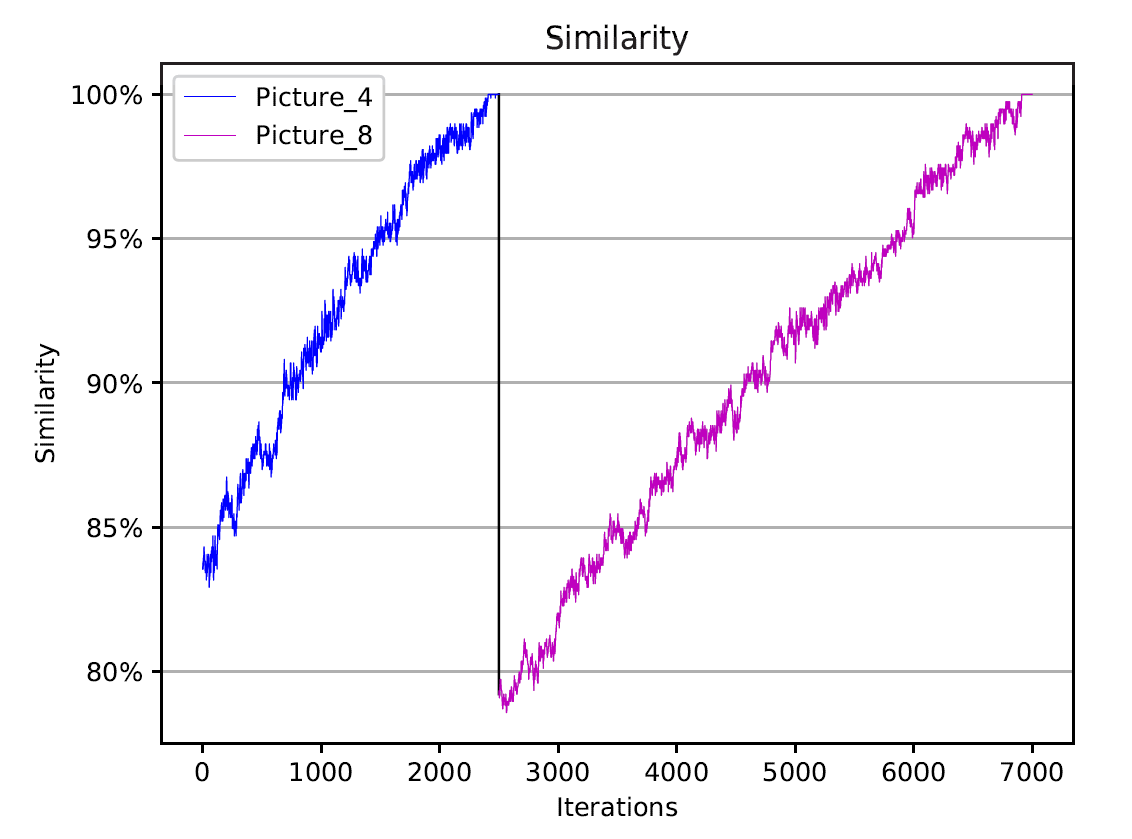}
		\setlength\abovecaptionskip{0pt}
		\setlength\belowcaptionskip{-0pt}
		\caption{The utility value during the stigmergy learning process.}
		\label{fig13}
	\end{center}
\end{figure}

The relevant simulation results are shown in Fig. 12. $n$ is the number of learning iterations. $\overline{d}$ represents the corresponding average distance from the others to the neural agent which should be excitatory at each iteration step. Each picture in Fig. 12 describes the current states of all neural agents during the learning process. White points in each picture indicate that neural agents at those locations are excitatory while black points indicate that neural agents at those locations are inhibitory. As the number of iterations increases, the learning system gradually converges to the clear target pattern of number 4 or 8. At the same time, the average distance $\overline{d}$ decreases gradually, indicating that the average amplitude of response of these neural agents increases gradually. After $n=2400$, the learning system starts to change its regulation and to express another target pattern of number 8. As before, the sitgmergy learning system finally learns the target pattern of number 8 after 7000 iterations. Fig. 13 shows the similarity between the original target picture and the one formed by the learning system during the whole process.

The above results have proved that the proposed stigmergy learning system can be adjusted to learn the target patterns, which is accomplished by activating the relevant sets of neural agents. Regardless of the initial stimulus, the activation of a neural agent is determined by the distance to other neuron individuals. Note that the distance represents the strength of interactions between neural agents which will be activated more easily with shorter distances from the others. In summary, the regulation of inter-synapse distance  plays an important role for the cooperation of neural agents in the stigmergy learning mechanism.

\vspace{-5pt}
\section{Conclusions}
\label{sec:conclusion}
Stigmergy phenomena are widely discovered in natural colonies and perform well through the way of collective collaboration. Inspired by the new discoveries on astrocytes in synaptic transmission, we have explored and mapped stigmergy in the regulation of synaptic activities in the brain. In particular, the interaction between neural agents  (synapses) is divided into three important phases and a stigmergic learning system model has been put forward. We have found that the regulation of distance between neural agents plays an important role in the proposed model. The well-regulated distance between neural agents can bring gain for the system and help to learn the target patterns. Its importance has been verified in two different simulations. Please note that the interaction between synapses within a certain range has been regarded as the short-range regulation. But for the long-range regulation, the participation of $\rm{IP_3}$ must be taken into account, which will be our future research direction.

\end{document}